\documentclass[sigconf, screen, final]{acmart}
\AtBeginDocument{%
	\providecommand\BibTeX{{%
			\normalfont B\kern-0.5em{\scshape i\kern-0.25em b}\kern-0.8em\TeX}}}

\copyrightyear{2021} 
\acmYear{2021} 
\setcopyright{acmlicensed}\acmConference[FDG'21]{The 16th International Conference on the Foundations of Digital Games (FDG) 2021}{August 3--6, 2021}{Montreal, QC, Canada}
\acmBooktitle{The 16th International Conference on the Foundations of Digital Games (FDG) 2021 (FDG'21), August 3--6, 2021, Montreal, QC, Canada}
\acmPrice{15.00}
\acmDOI{10.1145/3472538.3472540}
\acmISBN{978-1-4503-8422-3/21/08}

\usepackage[acronym, nomain]{glossaries}
\usepackage{graphicx}

\usepackage{multirow}
\usepackage{mathrsfs}

\usepackage[utf8]{inputenc}
\usepackage[T1]{fontenc}
\usepackage[english]{babel}
\newacronym{gan}{GAN}{Generative Adversarial Network}
\newacronym{nn}{NN}{neural network}
\newacronym{ml}{ML}{machine learning}
\newacronym{ea}{EA}{evolutionary algorithm}
\newacronym{es}{ES}{early stopping}
\newacronym{nlp}{NLP}{natural language processing}
\newacronym{cnn}{CNN}{convolutional neural network}
\newacronym{lstm}{LSTM}{Long Short-Term Memory}
\newacronym{ai}{AI}{Artificial Intelligence}
\newacronym{gpu}{GPU}{Graphics Processing Unit}
\begin{document}
\glsdisablehyper

\title{Dealing with Adversarial Player Strategies in the Neural Network Game iNNk through Ensemble Learning}

\author{Mathias Löwe}
\email{malw@itu.dk}
\orcid{0000-0003-1954-9653}
\affiliation{%
	\institution{IT University of Copenhagen}
	\city{Copenhagen}
	\country{Denmark}
}

\author{Jennifer Villareale}
\email{jmv85@drexel.edu}
\affiliation{%
	\institution{Drexel University}
	\city{Philadelphia}
	\state{Pennsylvania}
	\country{USA}
}

\author{Evan Freed}
\email{emf67@drexel.edu}
\affiliation{%
	\institution{Drexel University}
	\city{Philadelphia}
	\state{Pennsylvania}
	\country{USA}
}

\author{Aleksanteri Sladek}
\email{ams939@drexel.edu}
\affiliation{%
	\institution{Drexel University}
	\city{Philadelphia}
	\state{Pennsylvania}
	\country{USA}
}

\author{Jichen Zhu}
\email{jichen.zhu@gmail.com}
\orcid{0000-0001-6740-4550}
\affiliation{%
	\institution{Drexel University}
	\city{Philadelphia}
	\state{Pennsylvania}
	\country{USA}
}

\author{Sebastian Risi}
\email{sebr@itu.dk}
\orcid{0000-0003-3607-8400}
\affiliation{%
	\institution{IT University of Copenhagen}
	\city{Copenhagen}
	\country{Denmark}
}



\renewcommand{\shorttitle}{Dealing with Adversarial Player Strategies in the Neural Network Game iNNk}

\begin{abstract}
		Applying \gls{nn} methods in games can lead to various new and exciting game dynamics not previously possible. However, they also lead to new challenges such as  the lack of large, clean datasets, varying player skill levels, and 
		 changing gameplay strategies. In this paper,  we focus on the adversarial player strategy aspect in the game \emph{iNNk}, in which players try to communicate secret code words through drawings with the goal of not being deciphered by a \gls{nn}. Some strategies exploit weaknesses in the \gls{nn}  that
		   consistently trick it into making incorrect classifications, leading to unbalanced gameplay. We present a method that combines \emph{transfer learning} and \emph{ensemble methods} to obtain a data-efficient adaptation to these strategies. This combination significantly outperforms the baseline \gls{nn} across all adversarial player strategies despite only being trained on a limited set of adversarial examples.
		   We expect the methods developed in this paper to be useful for the rapidly growing field of \gls{nn}-based games, which will require new approaches to deal with unforeseen player creativity.

\end{abstract}

\maketitle

\renewcommand{\shortauthors}{Löwe et al.}

\glsresetall

\section{Introduction}
With the recent boom in \gls{nn} applications, game designers have been increasingly exploring a variety of \gls{nn} approaches in computer games~\cite{zhu2021player}. 
These include approaches where a \gls{nn} is directly incorporated into the gameplay experience or as a method for
dynamically generating content that would otherwise be created by a human artist~\cite{zhu2021player}. 
This approach has been utilized in well-known games such as \emph{Black and White}~\cite{wexler2002artificial}, \emph{Creatures}~\cite{creatures:grand1997creatures}, and \emph{Forza Motosport}~\cite{gitlin_2020}, which adapt game agent behavior in response to player input. In these cases, the \gls{nn} makes gameplay more personalized and potentially more engaging. 

With the recent advances, it is not surprising that more games utilize \glspl{nn} to provide novel gameplay experiences. While these novel experiences would not be possible without \glspl{nn}~\cite{cimolino2019oui,aidungeon}, designers are faced with new challenges when applying this technology. For instance, players may learn gameplay strategies that exploit weaknesses in the \gls{nn}, resulting in unintended interactions with the game. Adapting a \gls{nn} to account for this can be a challenging task, as effectively utilizing supervised learning with \glspl*{nn} often requires a large, labeled dataset. This is due to the considerable number of free parameters in the \gls{nn} architecture and the high dimensional input space. How well this dataset portraits the target behavior of the \gls{nn} has significant influence on the final performance. Such large, clean datasets which accurately represent the task are rare and expensive to obtain.

\begin{figure*}[tbh]
	\centering
	\includegraphics[width=1.\linewidth]{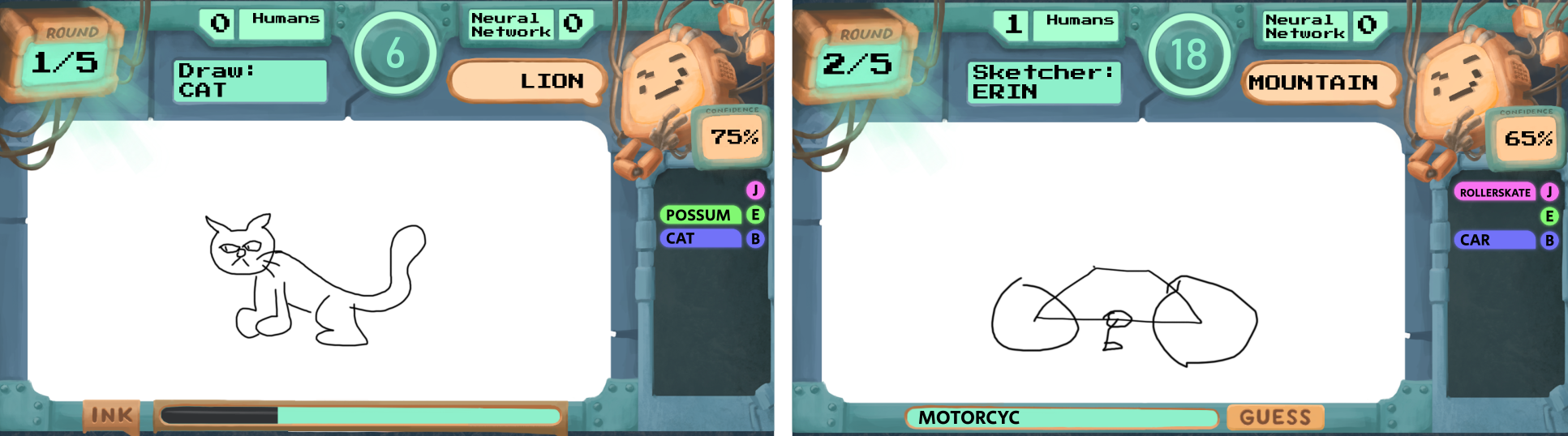}
	\caption{\textbf{The \emph{iNNk} user interface.} The user interface of the game \emph{iNNk} as presented in~\cite{villareale2020innk}. Left: The interface for the Sketcher who is tasked to draw a code word on the canvas. Right: The interface for the Guesser, who see the Sketcher's drawing with live updates. The Guesser's task is to decipher the sketch and correctly guess the code word before the NN.}
	\Description[The \emph{iNNk} user interface.]{The user interface of the game \emph{iNNk} as presented in~\cite{villareale2020innk}. Left: The interface for the Sketcher who is tasked to draw a code word on the canvas. Right: The interface for the Guesser, who sees the Sketcher's drawing with live updates. The Guesser's task is to decipher the sketch and correctly guess the code word before the NN.}
	\label{fig:innk_ui}
\end{figure*}

In this paper, we unpack the technical challenges that developers face when using a \gls{nn} in games and provide an approach to counter the player strategies that exploit the \gls{nn}. We utilize the \gls{nn}-based game, \emph{iNNk}~\cite{villareale2020innk}, as a case study to introduce a new \gls{nn} training procedure that leverages \emph{transfer learning} and \emph{ensembling} \gls{ml} techniques. \emph{iNNk}~\cite{villareale2020innk} is a web-based multiplayer drawing game where two or more people play together against a \gls{nn} (Figure~\ref{fig:innk_ui}).
To win the game, the players need to successfully communicate a secret code word to each other through drawings without the \gls{nn} deciphering the drawing. This game is particularly well suited for our purpose because the game mechanics encourage players to exploit the \gls{nn}. Understanding and preparing for such exploitative strategies is essential to balance gameplay.

Our method trains an ensemble consisting of the original \gls{nn} and a set of \gls{nn} specialists. Each specialist is trained exclusively on a specific player strategy that we observed overexploit the \gls{nn}. An example of such a player strategy is what we refer to as \emph{distraction}. Using this strategy, the player adds random strokes to the drawing canvas. These strokes work as noise and successfully stump the \gls{nn}. We use \emph{transfer learning} to clone the state of the original \gls{nn} to each specialist in order to kick-start the training process. Our method differs from existing \gls{ml} literature~\cite{breiman1996bagging, freund1996experiments, wolpert1992stacked, parmanto1996improving} as each ensemble specialist is built upon the same baseline \gls{nn} state with its individually distinct dataset to address a particular adversarial player strategy. Further, defense mechanisms for \glspl*{nn} against adversarial \emph{human-generated} data is a less explored field, and we demonstrate the efficiency of this approach for such settings with sparse datasets.

Our approach allows \emph{iNNk}'s \gls{nn} to successfully classify the sketches that utilize these exploitative strategies that the game designer may not have foreseen. A key benefit of this approach is that it requires only a limited number of labeled examples. We test our approach against the set of examples generated for each player strategy. In addition, we evaluate the impact our modifications have on the performance of the original task of classifying regular, non-adversarial sketches. Importantly for games, our method is computationally and temporally efficient, only requiring a limited amount of data. Lastly, it does not significantly compromise the model's performance on the original task, and the game is more prepared for game-breaking player strategies. 

With the increase of \gls{nn}-based games, how to address unforeseen player creativity and ensure a balanced gameplay is an exciting new research field. In this work, we present the steps in this direction with a \gls{nn} ensemble learning-based approach that enables learning from relatively sparse data and is generally applicable to a wide range of games. Our experiments show consistent performance improvements on all observed exploitative player strategies, without suffering catastrophic forgetting on the original task. These results highlight the potential of the approach presented in this paper.

\section{Background}
Numerous studies have examined a variety of \gls{nn} approaches used to introduce experimental features in existing games~\cite{Fontaine_2021, Awiszus_Schubert_Rosenhahn_2020, Volz_2018, Thakkar_2019}. We focus on games built with a \gls{nn} as an integrated part of the original game in this work. Particularly, we focus on how game developers have applied \glspl{nn} to solve a variety of tasks in their game. For instance, NNs have been used to procedurally generate a variety of different game content~\cite{gar:hastings2009evolving, risi2015petalz}. In the game \emph{GAR}~\cite{gar:hastings2009evolving}, the NN is responsible for generating new weapons for the player to use in battle. This content would otherwise be generated by a human artist, making the game more variable based on the player's preferences. \glspl*{nn} are also used to vary the behavior in non-player characters (NPCs). For example, in \emph{Creatures}~\cite{creatures:grand1997creatures}, players raise alien creatures, known as Norns, whose behavior is controlled by a \gls{nn}. When a new Norn is born, players direct the behaviors of the Norn through positive (petting), negative (striking), and text input to develop the actions and traits that will lead it to survive the longest.

Among \gls{nn}-based games, there are two types of learning methods: \emph{offline} and \emph{online learning}. In offline learning games, they use a fixed NN that is not adaptive during gameplay.
A notable example of offline learning is \emph{AI Dungeon}~\cite{aidungeon}, which is a text-based adventure game. Players are free to enter any text input to direct the story, and the \gls{nn} can still respond in reasonable ways. In the latest version of the game, \emph{Dragon}~\cite{ai_dungeon_dragon_2020}, the \gls{nn} used is commonly known as GPT--3. GPT--3 is trained on a dataset with $570$ GB of text using $175\text{B}$ parameters, and cost $\$12\text{M}$ to develop~\cite{brown2020language}. Rarely, game-developers have the time and resources to develop such massive \glspl*{nn} from scratch. Collecting such a large and clean dataset is unrealistic for most games. Further, due to the high dimensionality and lack of interpretability of modern \glspl*{nn}, making guarantees of their behavior is not feasible. Even well-performing \glspl*{nn} in offline learning games can easily be exploited by misleading them with adversarial input, as demonstrated in section \ref{sec:strategies}.

For online learning games, the \gls{nn} is continuously trained as the player interacts with it. Therefore, these games can adapt to individual players' actions in real-time. However, designers need to account for the possible ways players may exploit the training procedure, as this can quickly lead to an unbalanced gameplay experience. 

The general lack of guarantees on a \gls{nn}'s behavior makes \glspl{nn} hard to deploy in both online and offline settings. A sublime goal for using \glspl*{nn} in games is an automatic adaption to emerging player strategies without unintended side-effects. An example of such unintended side-effects is introducing strong predictive bias, e.g.\@ by always generating the same output regardless of the input. Another example is devoting substantial attribution to non-robust features, also known as overfitting. Therefore, to achieve automatic adaptation of emerging player strategies, numerous challenges need to be resolved that apply to online and offline learning games. These challenges include 1) sparsely collected and labeled datasets, 2) time-consuming training processes, and 3) the risk of overcompensating with catastrophic loss of performance on the original task solved by the \gls{nn}. 

In this paper, we aim to unpack the technical challenges and offer an approach to better account for and mitigate emerged player exploitations of the NN in the game \emph{iNNk}. First, we will review related literature on misleading \glspl{nn} and some related countermeasures. Then, we present our method with a case study on the game \emph{iNNk}. Finally, we present our results and discuss future possible improvements to the method.

\subsection{NNs Are Easily Fooled}
Misleading a \glspl{nn} is not an uncommon endeavor. Adversarial attacks against NNs are an established field of research in the \gls{ml} community. Research has repeatedly shown how NNs are easily fooled, and has demonstrated how to trigger NNs to make incorrect predictions with high confidence~\cite{nguyen2015deep, morgulis2019fooling, alcorn2019strike, Goodfellow2015ExplainingAH}. For instance, the concept of \glspl{gan} is founded on its ability to trick a \gls{nn} to believe an artificially generated image is real~\cite{goodfellow2014generative}. This problem has led to the creation of many interesting artifacts, from inferring missing information~\cite{yu2018generative} to the generation of completely new data~\cite{radford2016unsupervised, brock2019large}. Despite these seemingly innocent applications, the technique can also be used as an attack with severe consequences. An example is the modification of traffic signs such that a NN would misclassify a `Stop' sign for a `Speed Limit' sign~\cite{eykholt2018robust}. 
To fool a NN, a common approach is using gradient descent to generate adversarial examples~\cite{goodfellow2014generative, Goodfellow2015ExplainingAH, kurakin2017adversarial}. Others use \glspl*{ea}~\cite{Su_2019, ilyas2018blackbox} or human-generated data~\cite{eykholt2018robust}.

Recently, a new research area has emerged to counter these adversarial methods and provide alternative defensive solutions~\cite{guo2018countering, warr2019strengthening, lyu2015unified, ross2017improving, hinton2015distilling, papernot2016distillation}. This work has ignited an arms race in the field with continuous developments on both attack and defense mechanisms. Despite the numerous defense methods developed, they focus on computer-generated data to strengthen the robustness of the NN. However, what seems to have gained little attention thus far is developing defense mechanisms against adversarial attacks using human-generated data and make the NN adapt to such datasets. 

\subsection{Defending Against Adversarial Attacks with Sparse Datasets}
One approach to defend against adversarial attacks is the use of \emph{ensemble methods}. For instance, \citeauthor{abbasi2017robustness}~\cite{abbasi2017robustness} used a confusion matrix to choose the training dataset for fitting an ensemble of NN specialists. In general, ensemble methods have been shown to yield more robust models~\cite{hansen1990neural}. The baseline approach to ensemble methods involves having a set of ensemble members (weak learners), which, when combined, perform better than individual members alone~\cite{maclin2011popular}. This approach assumes the ensemble members are accurate and diverse~\cite{dietterich2000ensemble, hansen1990neural}. The ensemble approach can also leverage different types of weak learners and have proven successful with NNs~\cite{hansen1990neural}.

Alternate approaches to ensemble methods have a considerable variation in current literature~\cite{breiman1996bagging, freund1996experiments, wolpert1992stacked, parmanto1996improving}. 
Some methods, like boosting~\cite{freund1996experiments}, train each ensemble member sequentially. Each weights the importance of each entry in the training dataset by the previous model's loss in that example -- a higher loss gives a higher priority. Boosting decreases the ensemble's bias by focusing on ill-performing parts of the dataset. However, this also increases the risk of overfitting. Other ensemble methods, like bagging, have proven less prone to overfitting as they decrease the variance of the predictions. They are also effective with unstable learners due to the smoothing effect of model averaging
~\cite{dietterich2000ensemble, zhou2012ensemble}. To achieve a generalization improvement with bagging, diversity among the ensemble members is essential. This diversity results in a lower error correlation, as the ensemble members are less likely to make the same mistake on the same data~\cite{goodfellow2016deep}. 

Training robust, supervised, \gls{ml} models from the ground up usually requires a large and labeled dataset. If one does not have access to such a dataset, one can utilize the \gls{ml} technique called \emph{transfer learning}. This technique leverages the finding that using a fitted model on one dataset might serve as a good starting point for training a new model on a similar dataset~\cite{yosinski2014transferable}. Transfer learning is of particular use in domains with sparsely labeled datasets to help prevent overfitting~\cite{tan2018survey, wei_2017}. Further, it allows for utilizing fewer resources to acquire a model for a new task by adjusting the capabilities of an existing model~\cite{cs231}. Transfer learning has proven successful across a wide range of \gls{ml} problems, from \gls{nlp}~\cite{blitzer2007biographies, huang2013cross, mozafari2019bertbased} to image recognition~\cite{Yuqing_2018, Dawei_2019, 7404017}.

\section{Method}
\label{sec:method}
Our approach uses a \gls{nn} training procedure that leverages the \emph{transfer learning} and \emph{ensembling} \gls{ml} techniques. Our method trains an ensemble consisting of the original \gls{nn} and a set of \gls{nn} specialists. Each specialist is trained exclusively on a specific player strategy. We use \emph{transfer learning} to clone the state of the original \gls{nn} to each specialist to kick-start the training process.
To demonstrate our approach, we utilize the \gls{nn}-based game, \emph{iNNk}~\cite{villareale2020innk}. In this game, some of the emerged drawing strategies have shown to overexploit the \gls{nn} to an extent where players would always win. This outcome resulted in an unbalanced gameplay experience, as applying this strategy would consistently stump the NN regardless of the object to be drawn.

We aim to decrease the model's bias on the adversarial strategies while maintaining the best possible performance on the original dataset. The most straightforward approach would be to simply append these new instances to the existing dataset and continue the training process with this augmented dataset. This approach, however, might not yield the best player adaptation as discussed next. 

In case the model's capacity is already fully used, reducing bias on the adversarial examples would result in reduced performance on the original dataset. This happens because the adversarial examples are out-of-distribution samples, meaning each developed player strategy can potentially result in sketches looking substantially different from the ones in the original training dataset. Hence, the function the NN needs to fit, capturing both the original data \emph{and} the out-of-distribution samples, is more complex~\cite{goodfellow2016deep}. Without additional model capacity, the NN will have to `forget' some of the learned rules that applied to the original dataset, to learn rules applicable to the out-of-distribution data.

If the model's capacity is \emph{not} fully utilized, the NN could learn the features of the new dataset without losing performance on the old. However, for the case studied in this paper, the labeled dataset of an adversarial strategy is heavily underrepresented compared to the original training data. The sparsity of such collected datasets raises some difficulties as it would be easier for the NN to neglect this small fraction of outliers than to reduce the bias on these. Therefore, it is unlikely for such an approach to result in a \gls{nn} capable of accurately recognizing and accounting for the emerged player strategies.

Instead, inspired by \emph{ensemble methods} and \emph{transfer learning}, we propose a combination with an ensemble of model specialists. We use the same NN structure for all ensemble members, as detailed in section \ref{sec:nn_setup}. First, we fit a model to the original, non-adversarial dataset. We will refer to this as our baseline model. Then, we collect and label the player data. Each example of a player strategy gets stored in a dataset with instances of only that player strategy.

\begin{figure*}[tb]
	\centering
	\includegraphics[width=1.\linewidth]{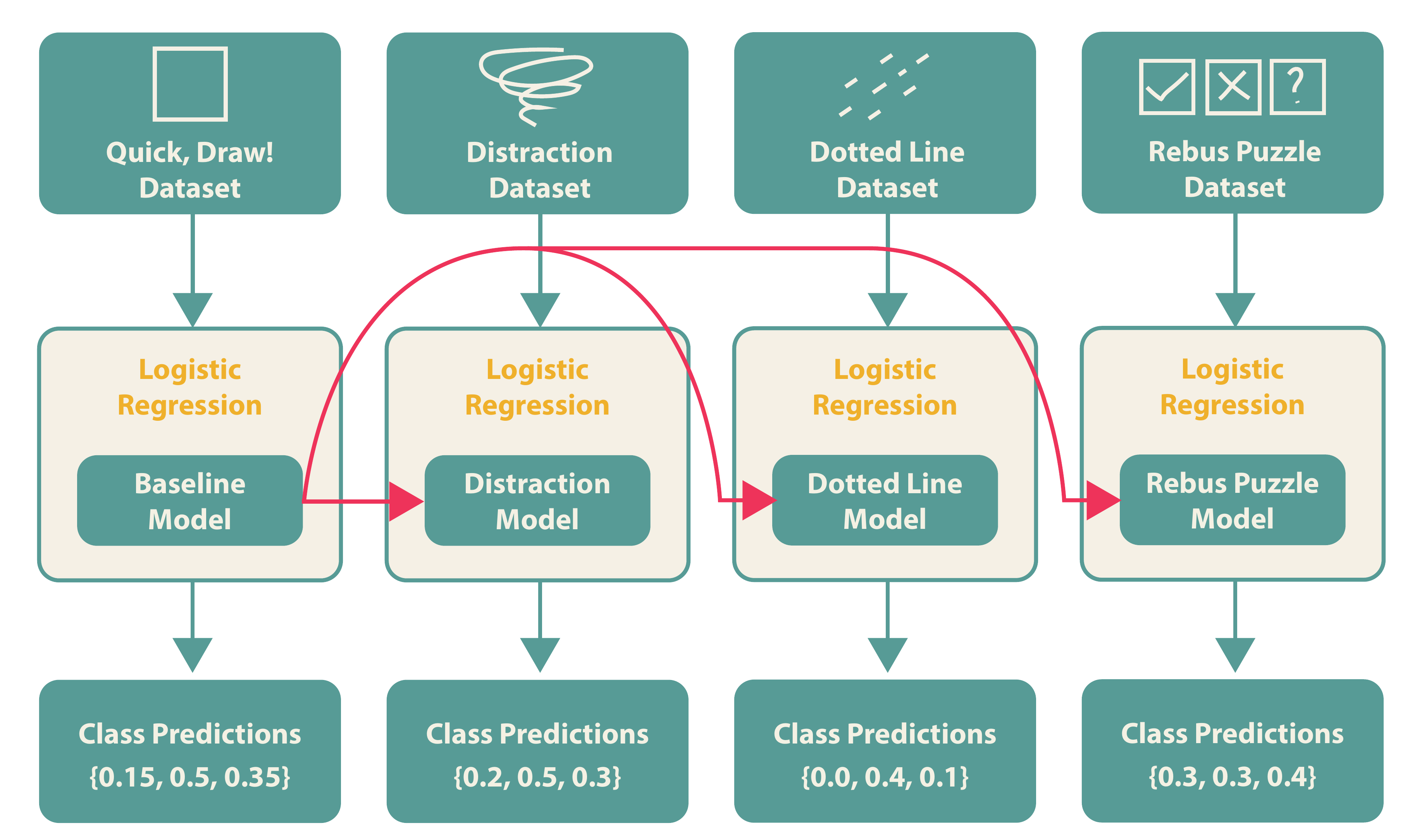}
	\caption{\textbf{An overview of the ensemble training procedure.} Before training a model, the state of it requires initialization. For this, we use transfer learning. A red arrow indicates the initialization of a model's state. The state of the model at the arrow's end is initialized to the state of the model at the origin of the arrow. Each model is assigned its own, distinct, dataset, and is specialized in that particular set. All models are trained using logistic regression for the classification of stroke data to one of 345 classes.}
	\Description[]{}
	\label{fig:ensemble_training}
\end{figure*}

As the examples generated using each of the player strategies are similar to the original dataset, we use transfer learning to circumvent severe overfitting and minimize the cost of adapting to each of these adversarial strategies.
The training process is initialized by transferring the state of the baseline NN to the ensemble specialist.
Each ensemble member is trained \emph{independently} of the others on its designated dataset. An overview of the training procedure is shown in Figure~\ref{fig:ensemble_training}.

\begin{figure*}[tbh]
	\centering
	\includegraphics[width=.75\linewidth]{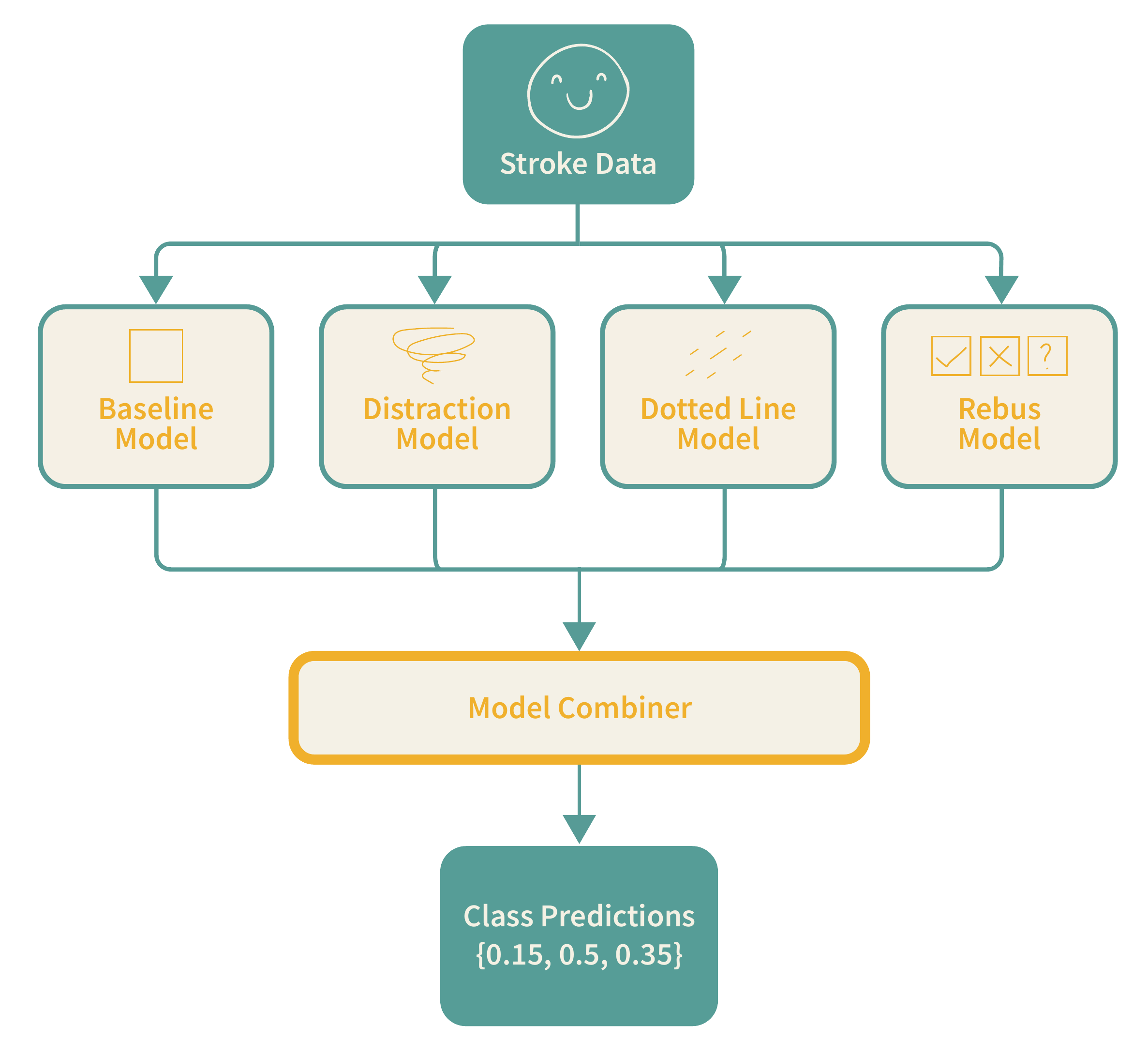}
	\caption{\textbf{Making predictions with the ensemble.} To make predictions, we employ all ensemble members and query each for a prediction based on the same observation. The set of model predictions needs to be combined into a single prediction to compute the final, class-wise, probabilities.}
	\Description[]{}
	\label{fig:ensemble_predictions}
\end{figure*}

After the training phase, all models are combined into an ensemble, as illustrated in Figure~\ref{fig:ensemble_predictions}. For making predictions with the ensemble, each member is queried for a prediction given the same observation. The resulting set of predictions needs to be combined into a single vector of class-wise probabilities. While any combinatoric method can be applied, we employed plain model averaging in our experiments. 

\section{Case Study: iNNk} \label{sec:innk}
To test our method, we use the game \emph{iNNk} as a case study. The game is a web-based multiplayer drawing game where two or more people play together against a \gls{nn}. To win the game, the players need to successfully communicate a secret code word to each other through drawings without being deciphered by the \gls{nn}. Players are assigned one of the two roles during the game: the Sketcher and the Guesser. The Sketcher is tasked with drawing something based on the code word assigned by the game. The goal is to draw the code word so that the human Guesser can interpret the code word accurately before the NN. The Guessers are tasked with entering their guess of the code word based on the Sketcher's drawing before the NN guesses correctly. The NN always plays the role of a Guesser, and its goal is to decipher correctly first. Screenshots of the game are shown in Figure~\ref{fig:innk_ui}. 

\subsection{Adversarial Player Strategies} \label{sec:strategies}
\begin{figure*}[tbh]
	\centering
	\includegraphics[width=1.\linewidth]{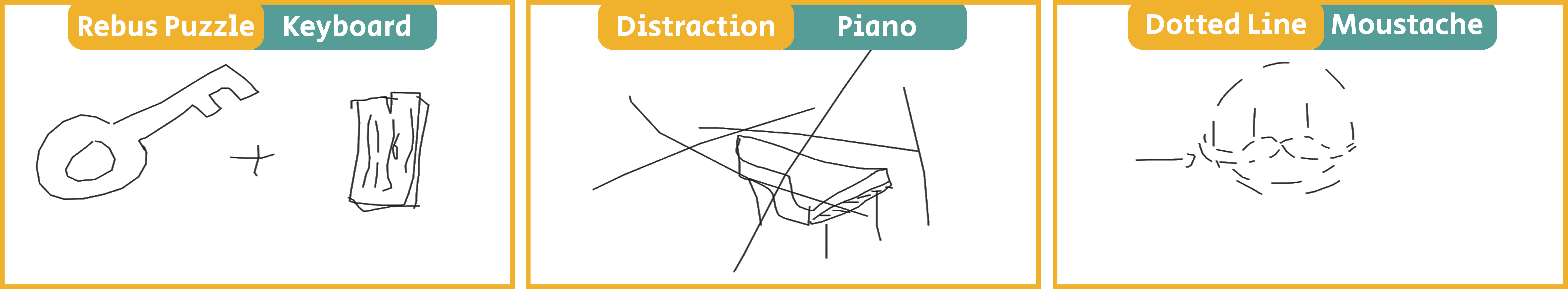}
	\caption{\textbf{Examples of the three adversarial player strategy sketches used for retraining: Rebus Puzzle (left), Distraction (center), and Dotted Line (right).} For the example of the Rebus Puzzle strategy, the code word is `Keyboard'. The Sketcher decided to divide this into two separate sketches, one of a key and one of a wooden board. An example of the Distraction strategy is a drawing for the code word `Piano'. Using this strategy, the Sketcher added straight lines to stump the NN. The superimposed sketch of a piano is otherwise unchanged. Finally, for the Dotted Line strategy, the example shown is for the code word `Moustache'. Here, the Sketcher only modifies the linestyle.}
	\Description[]{} 
	\label{fig:adversarial}
\end{figure*}

Investigating winning strategies employed by the players in  \emph{iNNk}~\cite{villareale2020innk} (i.e.\@, observing about eight players play iNNk over four games each),  
we found three different adversarial strategies that were commonly developed (Figure~\ref{fig:adversarial}).
These strategies are not exhaustive, and a more thorough user study will likely reveal many more strategies. Our objective is to demonstrate the technical challenges when dealing with emerging player strategies and provide an approach to alleviating these explorations in a \gls{nn}-based game for which these three initial strategies were deemed sufficient.

The first strategy includes the Sketcher drawing the code word in a sequential set of images (i.e.\@, as a rebus). The Sketcher was given the code word `eyeglasses', they sketched two separate images, an eye and a pair of drinking glasses, in an attempt to stump the NN. 
The second strategy includes adding visual noise or other shapes in addition to the drawing of the code word. In this case, the player would crosshatch or draw extra lines to mislead the NN. The players added these extra lines at the beginning of the round. Adding them later would allow the \gls{nn} to classify the drawing correctly before adding the distraction lines. The third strategy includes drawing the code word more stylistic (i.e.\@, using a dotted line instead of a solid line).

Based on these three observed adversarial player strategies, the game designers manually drew around one hundred sketches for each one ($105$ for Distraction, $93$ for Dotted, and $66$ for Rebus). This process was followed to create a clean dataset of each strategy with a sufficient size while minimizing labeling cost. For the Distraction strategy, this was as simple as appending arbitrarily placed lines on some part of the drawing canvas at the beginning of the sketch. It was similarly straightforward to generate data for the Dotted Line strategy by drawing while releasing the mouse periodically. The Rebus Puzzle strategy presented some issues as not all of the classes in our dataset lend themselves to this approach. For example, as seen in Figure \ref{fig:adversarial}, the class `keyboard' can be broken into a key and a board of wood which a human player may understand while the NN cannot. However, if the class chosen by the game is `dog', the word cannot be broken down using this strategy at all. This resulted in a sparser dataset with a less diverse set of classes for training a model on this strategy.

\subsection{Neural Network Architecture and Training Details}
\label{sec:nn_setup}
The task of the NN is to classify a sketch correctly given a sequence of stroke data. 
As mentioned in Section~\ref{sec:method}, each NN in the ensemble shares the same structure and is similar to the one introduced in a tutorial by~\citeauthor{tensorflow_tutorial}~\cite{tensorflow_tutorial}. 
Each of the ensemble NNs is first pre-trained on the dataset collected through Google's `Quick, Draw!' online game~\cite{quick_draw}. 
The dataset consists of approximately $40 \text{M}$ sketches of 345 distinct classes. Our experiments utilized $10 \%$ of it as a validation dataset and $20 \%$ as a testing dataset. The sketches in this dataset are human-made digital drawings represented as temporally ordered sequences of \emph{strokes} that make up the drawing. Each \emph{stroke} in this sequence is a continuous curve. The curve starts when the sketcher's drawing tool (such as a computer mouse) was activated and ends when the drawing tool was deactivated. This curve is represented as a sample of temporally ordered 2-D point coordinates along the curve. In addition to the recording of the sketch, the dataset also contains metadata for each sketch, the most relevant of which for training our NN was the label of what each sketch depicts. We utilized the `simplified' version of this dataset in our training process, in which the sketches had been preprocessed by centering and scaling them into a $256 \times 256$ space. The strokes had also been simplified with the \emph{Ramer-Douglas-Peucker} algorithm~\cite{RAMER1972244}. We also applied further preprocessing in the NN training process in the form of min-max normalization of the strokes. In addition to the temporally ordered 2-D coordinates, the NN also receives a binary signal, indicating when a stroke is ended. Due to the small datasets containing the user strategies, we used a testing and validation dataset of $10\%$ for these.

Each \gls{nn} contains convolutional, recurrent, and dense layers. First, a series of three 1-D convolutional layers map the input into $\left[48, 64, 96 \right]$ feature maps using kernel sizes $\left[5, 5, 3 \right]$. The first convolutional layer takes three input channels due to the nature of the data. After each layer, dropout~\cite{srivastava2014dropout} and batch normalization~\cite{ioffe2015batch} are applied to regularize the network. The feature maps are then consumed by the recurrent layers. These layers are responsible for memorizing features across time. We use three bi-directional \gls{lstm} layers\cite{lstm_paper}, each with a hidden state of size $256$. After each \gls{lstm} layer, we also apply a dropout layer. Finally, a dense, fully connected layer computes a class-wise list of classification likelihoods. This layer takes an input vector with a cardinality of $512$ due to the bi-directional \gls{lstm} layers. Because of the multinomial logistic regression, the output is a vector of logits. These are modified using the Softmax function to convert them into probabilities over the $345$ distinct classes. During training, we compute the cross entropy loss between these predictions and the ground-true label.

The hyperparameters of the training process remain unchanged for all ensemble members. Namely, we apply early stopping~\cite{prechelt1998early} with a patience of $20$ for the training of all ensemble members, including the baseline model. We use a learning rate of $3 \times 10^{-4}$, a batch size of $256$, the adam optimizer for gradient descent~\cite{kingma2014adam}, a dropout~\cite{srivastava2014dropout} rate of 0.3 for all dropout layers, and Xavier initialization~\cite{pmlr-v9-glorot10a} for the free parameters of the baseline model. To prevent exploding gradients of the recurrent model, we apply gradient clipping
~\cite{goodfellow2016deep} with a maximum $\ell ^2\text{-norm}$ of $1$. The experiments were executed on a machine with an Intel Core i7-5820K CPU, with $64\text{GB}$ of RAM and an NVIDIA GP102 TITAN X GPU.

\section{Results}
\begin{table*}[htbp]
	\caption{Performance on the various datasets for each model on a set of performance metrics.}
	\label{tab:results}
	\resizebox{\textwidth}{!}{%
		\begin{tabular}{|l||l|l|l|l||l|l|l|l||l|l|l|l|}
			\hline \multirow{4}{*}{\textbf{Neural Network}} &
			\multicolumn{12}{c|}{
				\textbf{Dataset}} \\\cline{2-13} &
			\multicolumn{4}{c||}{{\color[HTML]{569d97} Top-1 Accuracy}} &
			\multicolumn{4}{c||}{{\color[HTML]{569d97} Top-5 Accuracy}} &
			\multicolumn{4}{c|}{{\color[HTML]{569d97} Cross entropy}} \\ \cline{2-13}
			\textbf{} &
			\textbf{\begin{tabular}[c]{@{}l@{}}Quick,\\ Draw!\end{tabular}} &
			\textbf{Distraction} &
			\textbf{Dotted} &
			\textbf{Rebus} &
			\textbf{\begin{tabular}[c]{@{}l@{}}Quick,\\ Draw!\end{tabular}} &
			\textbf{Distraction} &
			\textbf{Dotted} &
			\textbf{Rebus} &
			\textbf{\begin{tabular}[c]{@{}l@{}}Quick,\\ Draw!\end{tabular}} &
			\textbf{Distraction} &
			\textbf{Dotted} &
			\textbf{Rebus} \\ \hline
			\textbf{\begin{tabular}[c]{@{}l@{}}Baseline\\ Model\end{tabular}} &
			$\mathbf{83.52 \%}$ &
			$0.00 \%$ &
			$0.00 \%$ &
			$0.00 \%$ &
			$\mathbf{96.20 \%}$ &
			$20.00 \%$ &
			$0.00 \%$ &
			$0.00 \%$ &
			$\mathbf{0.62}$ &
			$5.22$ &
			$7.52$ &
			$6.97$ \\ \hline
			\textbf{\begin{tabular}[c]{@{}l@{}}Distraction\\ Model\end{tabular}} &
			$80.51 \%$ &
			$\mathbf{70.00 \%}$ &
			$0.00 \%$ &
			$16.67 \%$ &
			$95.35 \%$ &
			$\mathbf{86.67 \%}$ &
			$25.92 \%$ &
			$16.67 \%$ &
			$0.74$ &
			$\mathbf{1.82}$ &
			$5.39$ &
			$11.46$ \\ \hline
			\textbf{\begin{tabular}[c]{@{}l@{}}Dotted\\ Model\end{tabular}} &
			$77.80 \%$ &
			$10.00 \%$ &
			$\mathbf{70.37 \%}$ &
			$0.00 \%$ &
			$94.33 \%$ &
			$30.00 \%$ &
			$\mathbf{88.89 \%}$ &
			$16.67 \%$ &
			$0.84$ &
			$6.00$ &
			$\mathbf{1.69}$ &
			$7.50$ \\ \hline
			\textbf{\begin{tabular}[c]{@{}l@{}}Rebus\\ Model\end{tabular}} &
			$79.32 \%$ &
			$13.33 \%$ &
			$7.41 \%$ &
			$\mathbf{11.11 \%}$ &
			$95.05 \%$ &
			$50.00 \%$ &
			$14.81 \%$ &
			$\mathbf{44.44 \%}$ &
			$0.77$ &
			$4.45$ &
			$5.76$ &
			$\mathbf{4.28}$ \\ \hline
			\textbf{\begin{tabular}[c]{@{}l@{}}Ensemble\\ Model\end{tabular}} &
			$82.59 \%$ &
			$50.00 \%$ &
			$0.00 \%$ &
			$16.67 \%$ &
			$95.98 \%$ &
			$70.00 \%$ &
			$44.29 \%$ &
			$33.33 \%$ &
			$0.65$ &
			$2.51$ &
			$4.42$ &
			$4.98$ \\ \hline
		\end{tabular}%
	}
\end{table*}

To test the performance of our proposed ensemble model with transfer learning, we evaluate each ensemble member and the ensemble itself on each dataset. Fitting the baseline model took $\approx22$ days, while fitting each ensemble specialist took ${<}1$ minute. The main objective is to maintain consistent performance across all datasets without losing performance on the original, non-adversarial, `Quick, Draw!' dataset. Maintaining a good performance on the original dataset is of high importance such that an entertaining gameplay experience for players with non-adversarial strategies can endure. The results are outlined in Table~\ref{tab:results}, and are averaged over three repeated experiments.

We evaluated each model on each dataset on three performance metrics: Top-1 Accuracy, Top-5 Accuracy, and Cross entropy loss. With the small testing dataset from each of the adversarial strategies at hand, the cross entropy loss is the better performance metric since accuracy is not a smooth measure. A decrease in cross entropy is often correlated with increased accuracy, but this is not always the case. With small datasets, minor modifications to a single prediction can yield high variations in the measurement accuracy. We included the accuracy metrics simply as these are more intuitive, but the cross entropy better describes discrepancies in the model's output distribution and the ground true target distribution. A higher cross entropy can indicate a less confident prediction, or indicate a confident, incorrect prediction. Finally, one should be careful comparing cross entropy loss across tasks, and the reported cross entropy losses should only be compared to other models on the same dataset. An example of considerable discrepancies between the cross entropy loss and the Top-1 Accuracy is the Distraction model's performance on the Rebus dataset. In this case, the model obtains a $16.67\%$ Top-1 Accuracy, which is higher than the Rebus model (on average), but this with a stunning $11.46$ cross entropy loss compared to the $4.28$ of the Rebus model. 

As shown in the table, each ensemble member performs the best on the dataset for which it is a specialist (the diagonal marked in bold). Yet, when used as stand-alone, each ensemble member generally does not perform well on the other datasets and suffer a loss of at least three percentage point on the `Quick, Draw!' dataset compared to the Top-1 Accuracy of the baseline model. The worst-case is the Dotted specialist with a performance loss of over five percentage points on the original task.

Further, it is clear from the table that the baseline model performs inadequately across all adversarial datasets. The only model performing consistently across all presented datasets is the ensemble model. While it suffers a performance loss on the Dotted dataset compared to the Dotted specialist, its cross entropy loss for this dataset is still highly improved over any other model. This is another example of disparities between the cross entropy loss and the Top-1 Accuracy. The Top-1 Accuracy suggests the ensemble model has made no improvement on the Dotted dataset. Yet, significant improvements are seen in the cross entropy, and these are reflected in the Top-5 Accuracy as well. Overall, the ensemble makes significant improvements on all adversarial datasets with a minimal loss in performance on the `Quick, Draw!' dataset across all performance metrics.

\section{Discussion}
\label{sec:discussion}

The Dotted model's loss on the original dataset could indicate that the parameters of the model need substantial adjustments for improving performance on the Dotted dataset. One explanation could be the low-level feature maps in the first convolutional layers, which typically work as edge- and curve-detectors. These might transfer poorly from the `Quick, Draw!' to the Dotted dataset, as an edge in the Dotted dataset no longer indicates separation but is most often part of a continuous line.

The motivation behind reporting the Top-5 Accuracy as a performance metric is due to the nature of the game \emph{iNNk}. Here, an incorrect guess gets blacklisted, and neither players nor NN is allowed to make this prediction again. Technically, this happens by setting the NNs predicted logits to $-\infty$ for these blacklisted classes. For instance, when the class with the highest probability is blacklisted, the NN would predict its perceived second most likely class. This is to prevent the repeating of incorrect guesses. To prevent exhausting the list of whitelisted classes, the NN is only queried for a prediction every 2.5 seconds. For this reason, having the correct class among the Top-5 predictions is important for the gameplay, as the four incorrect guesses would be blacklisted within 10 seconds at worst. One could argue that not having a perfect Top-1 accuracy score even improves the gameplay experience, as an oracle-like NN might discourage players from playing the game. From the results in Table~\ref{tab:results}, it is clear that the proposed ensemble model gains significant improvements for all datasets on this Top-5 Accuracy.

An inherent limitation in our approach on how to prepare for adversarial player strategies is that to identify one, it must have a recognizable pattern that the \gls{nn} can be trained on. Therefore, our approach will not work on strategies developed when players communicate outside of the game to achieve the correct answer. This information cannot be made available to the \gls{nn}, meaning there is no way to compensate for this type of strategy. Other strategies may involve inside knowledge within the player group, which would also be challenging and most likely not particularly useful to train the \gls{nn} on. 

Additionally, the proposed process of identifying and collecting examples of player strategies is limited in terms of scalability. For instance, our work identified player strategies through human observations and analyses. However, identifying strategies on a much larger scale or using a more complex game with this process may not be as easily executed. To address these limitations, we plan on investigating the automation of these processes (see section \ref{sec:future}).

Finally, each time a \gls{nn} is added as a member of the ensemble, it prolongs the time it takes to make predictions. Making predictions with the ensemble can be run in parallel on separate GPUs. In practice, however, there is a limit to the scalability of the method and hence the number of addressable, distinct player strategies.

\section{Future Work}
\label{sec:future}
Our planned next steps involve identifying further player strategies and extending \emph{iNNk's} \gls{nn} to recognize them. Given that there could exist a large number of player strategies that dynamically emerge over the game's life-cycle, it would be beneficial to automate the process of discovering new player strategies and the retraining of the \gls{nn}. This would lead to a more dynamic experience of the game evolving over time and also reduce the amount of work in achieving this. Therefore, we plan to explore methods for automating the player strategy discovery process using \emph{clustering} \gls{ml} methods, for example. Given the difficulties with generating player strategy data by hand, we also plan to explore automated approaches for augmenting training data to produce new examples, such as via programmatic generation of player strategy drawings. Finally, we plan to perform further fine-tuning on our proposed ensemble via experimentation with different model prediction aggregation (voting) methods.

An additional point of investigation could be to assess the impact of batch normalization. Recent studies suggest the use of batch normalization can increase a \gls{nn}'s vulnerability to adversarial attacks~\cite{benz2020batch, galloway2019batch}. While more analysis is needed on this subject, future work could explore whether the baseline model applied in this research could benefit from abandoning the use of batch normalization in the model architecture. While such modification potentially increases model robustness, it will also remove the benefits from applying batch normalization -- faster model convergence being one of them~\cite{pmlr-v37-ioffe15}. This could lead to an unacceptable increased cost in model training, as the current model already takes $>\!3$ weeks to fit.

\section{Conclusion}
\label{sec:conclusion}
In this paper, we use the \gls{nn}-based game \emph{iNNk} as a case study to demonstrate an approach to adapt a NN for adversarial player strategies in order to maintain balanced gameplay. 
To counter those strategies, we propose an approach that combines transfer learning and ensemble methods to strengthen our classifier based on sparse datasets. We evaluate the performance of our approach using three different performance measures on the testing datasets. As a result, we found that each model dedicated to a strategy lost performance on the baseline dataset, while the baseline model was ineffective on all player strategies. We found our ensemble approach to provide the best, most consistent performance across all datasets, with $<\!\!1\%$ loss on Top-1 Accuracy on the original dataset compared to the baseline model.
Further, we found our use of transfer learning crucial for the ensemble setup to alleviate overfitting.
Since our method is effective and efficient to set up, despite our ensemble of models being trained on a very limited amount of data, it can be used by others to create more competent NNs in their own domains. Lastly, we present potential ways to develop this method further that include automating the process and building more sophisticated ensembles, which would result in more accessible and more interesting use cases for NNs in games.

\begin{acks}
	This project was partially supported by the Innovation Fund Denmark (Grant ID 8053-00073B), a DFF-Research Project1 grant (9131-
00042B), and U.S. National Science Foundation under grants IIS-1917855 and IIS-1816470. 
\end{acks}

\bibliographystyle{ACM-Reference-Format}
\bibliography{references}

\end{document}